\documentclass[conference]{IEEEtran}
\IEEEoverridecommandlockouts
\usepackage{cite}
\usepackage{amsmath,amssymb,amsfonts}
\usepackage{algorithm}
\usepackage{algorithmic}
\usepackage{graphicx}
\usepackage{textcomp}
\usepackage{xcolor}
\usepackage{soul}
\usepackage{todonotes}
\usepackage{multirow}
\usepackage{booktabs}
\usepackage{url}
\usepackage[T1]{fontenc}

\newcommand{\modelname}{\texttt{AQUAVS}}

\def\BibTeX{{\rm B\kern-.05em{\sc i\kern-.025em b}\kern-.08em
    T\kern-.1667em\lower.7ex\hbox{E}\kern-.125emX}}
\begin{document}

\bibliographystyle{IEEEtran}

\title{Assessing the Quality of the Datasets by Identifying Mislabeled Samples}



\author{
    Vaibhav Pulastya, 
	Gaurav Nuti,
	Yash Kumar Atri,
	Tanmoy Chakraborty
     \\
	IIIT-Delhi, New Delhi, India \\
	{\{vaibhav17271, gaurav17050, yashk, tanmoy\}@iiitd.ac.in}
}

\maketitle

\begin{abstract}
Due to the over-emphasize of the quantity of data, the data {\em quality} has often been overlooked. However, not all training data points contribute equally to learning. In particular, if mislabeled, it might actively damage the performance of the model and the ability to generalize out of distribution, as the model might end up learning spurious artifacts present in the dataset. This problem gets compounded by the prevalence of heavily parameterized and complex deep neural networks, which can, with their high capacity, end up memorizing the noise present in the dataset. This paper proposes a novel statistic --  \textit{noise score}, as a measure for the quality of each data point to identify such mislabeled samples based on the variations in the latent space representation. In our work, we use the representations derived by the inference network of data quality supervised variational autoencoder (\modelname). Our method leverages the fact that samples belonging to the same class will have similar latent  representations. Therefore, by identifying the outliers in the latent space, we can find the mislabeled samples. We validate our proposed statistic through experimentation by corrupting MNIST, FashionMNIST, and CIFAR10/100 datasets in different noise settings for the task of identifying mislabelled samples. We further show significant improvements in accuracy for the classification task for each dataset.
\end{abstract}


\section{Introduction} 

Deep learning models have seen great practical success in achieving state-of-the-art performance for various tasks. However, such models generally are heavy, with a large number of \textit{trainable} (or learnable) parameters, and are intrinsically constrained by the data they are trained on. This often leads to one of the major concerns for such deep learning models as due to their high capacity, they often overfit the training data provided to them irrespective of any artefacts that the data might contain \cite{arpit2017closer}. Therefore, if the training data is noisy (i.e., consisting of mislabeled data points), this results in hampering the model's generalizability. Additionally, due to the demand for large-scale datasets followed by the common dictum that more labeled data results in better performance,  dataset creation is generally automated, emphasising the {\em quantity} of the data rather than the {\em quality} of data. This increases the scope for noise in datasets, as in many settings, they might contain data that is "weakly labeled" through web-scraping or other similar techniques \cite{li2017webvision}. Although the data might be inexpensive to collect, these methods often lead to errors in labeling; thereby, training on such data instead ends up hampering the model performance. Furthermore, crowdsourcing platforms, which employ human annotators or rely on community feedback or survey for data labeling are also susceptible to making mistakes  \cite{10.1145/2858036.2858115}. Hence, in a real-world setting, it becomes inevitable to have a source of noise for large datasets.

This paper aims to automatically identify such mislabelled or ``hard'' data points without any availability of all ``clean'' a priori subset of data or any knowledge about the type of noise present in the dataset. Upon identification and filtration of such samples, we expect the training on the derived clean subset to improve the generalization of the model. We validate this by conducting a series of experiments on different datasets and against two different noise settings - random and systematic. A random sample of a class is mislabeled uniformly to any other class in case of random noise. Whereas in systematic noise, samples belonging to a specific class are mislabeled to another fixed class, which intends to imitate a scenario wherein samples belonging to a class may be confused for some other class during the annotation process.

\begin{figure*}[h]
    \centering
    \includegraphics[ width=20cm,
  height=4cm,
  keepaspectratio]{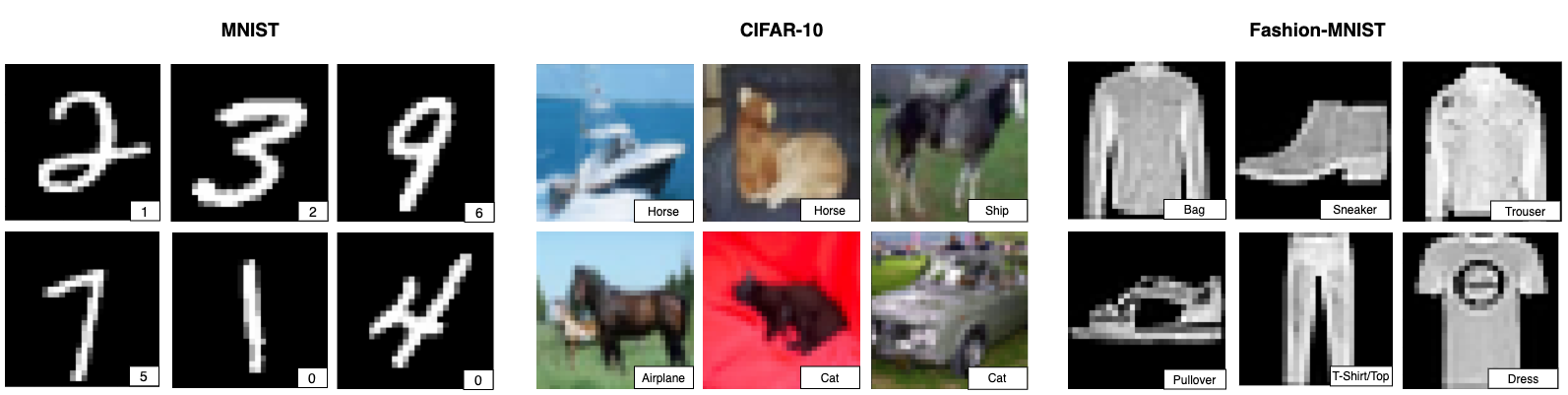}
    \caption{Examples of images and their corresponding assigned labels from the three datasets -- MNIST, FashionMNIST, and CIFAR-10, having high \textit{noise score} values. Such points are classified and expected to be mislabeled.
    \label{fig:highNoisesample}}
\end{figure*}

To combat the mentioned problem, we quantify the quality of each data point relative to the whole dataset. We introduce \modelname, a variational autoencoder based architecture with an  auxiliary network. \modelname\  incorporates data labels to effectively learn the latent representations of input data. We propose a novel method that utilizes the variations in latent representations. Our primary intuition is that latent space representations of samples belonging to the same class will be similar. Therefore, our metric measures how far off (outlying) is the latent representation of a sample to that of samples belonging to the same class. We define \textit{noise score} given by the number of outlying latent features of a sample.  We discuss our proposed approach in detail in Section \ref{methods}. As illustrated in Figure \ref{fig:highNoisesample}, we expect points with a high noise score to be mislabeled and apply threshold to filter out the resultant clean subset.

Lastly, we validate our method by conducting extensive experiments on four different datasets, namely MNIST \cite{lecun-mnisthandwrittendigit-2010}, FashionMNIST \cite{xiao2017fashionmnist}, CIFAR-10 \cite{Krizhevsky09learningmultiple}, and CIFAR-100 \cite{Krizhevsky09learningmultiple}. We assume the original datasets to contain all clean samples, corrupt them for our experiments using different noise models, and then report the results for each dataset. For our baseline, we discuss an approach that uses training dynamics of entropy of data sample. Recent empirical studies \cite{arora2019finegrained,swayamdipta2020dataset} suggested how training dynamics contain information that can be used to differentiate between samples vis-a-vis their contribution to learning.


The main contributions of this paper are summarized below:
\begin{itemize}
\item  We propose a novel algorithm that makes use of latent space representation of an input observation and returns a  quality measure (or noise score) for it. We use this to classify high noise score samples as mislabelled or hard-to-learn data points. 
\item Our method identifies noisy samples without any access to a clean subsample for deduction, or any prior knowledge on the type of noise present in the dataset. 
\item We conduct extensive experimentation for the task of identifying mislabelled points under different  noise settings -- uniform and systematic noise, on multiple well-known datasets, wherein our method shows significant improvement over baselines.
\end{itemize}

{\bf Reproducibility:} We have made the code and datasets public at \url{https://github.com/LCS2-IIITD/AQUAVS}.
\section{Related Works} \label{relatedwork}

The challenges with learning a noisy dataset have been well-explored  \cite{10.5555/2969033.2969226,6685834,Ekambaram}. A detailed survey of existing literature can be found in \cite{AlganIlkay,Song2020LearningFN}. To counteract noisy labels, various approaches such as forward or backward correction \cite{patrini2017making} and S-model \cite{DBLP:conf/iclr/GoldbergerB17}, suggest estimation of the noise distribution in terms of noise transition matrices for classes and further rectify the loss function based on it. The primary challenge is to accurately estimate the aforementioned matrices. Sabzevari et al. \cite{SABZEVARI20182374} identified the noise samples by training a classifier on subsampled data and utilizing the threshold to filter the mislabeled data points. Another technique applied in \cite{reed2015training} and \cite{ma2018dimensionalitydriven} involves rectifying the said noise by using the prediction of deep neural networks (DNN). However, this method poses a challenge of overfitting. To account for the issue of overfitting, in joint optimization, Tanaka et al. \cite{tanaka2018joint} introduced a regularization term that takes prior knowledge about noise into account, which is not always readily available in real-world settings. Another approach that has gained traction identifies samples that are important to training. This involves training models selectively on such samples or to weight samples according to the importance measure (e.g. \cite{DBLP:journals/corr/abs-1712-05055,DBLP:journals/corr/abs-1803-09050,article}). For this, the primary challenge is to design robust and compelling criteria to score the training samples.

One of the recent approaches, Labelfix \cite{Muller_2019} proposes such criterion given by the probability of the assigned original label for an input. Similarly, our proposed method does segregation by describing a novel measure -- \textit{noise score} as the standard for selectively filtering out the clean data points. The proposed pipeline follows a structure as the one by Brodley and Friedl \cite{Brodley_1999}. The two-step pipeline includes (i) identifying mislabeled samples and (ii) training models on the derived subset after discarding mislabeled samples. Additionally, many deep learning methods have been proposed to identify mislabeled samples (at times implicitly). These include Iterative Noisy Cross-Validation (INCV) \cite{chen2019understanding}, which uses cross-validation; MentorNet \cite{DBLP:journals/corr/abs-1712-05055} implements auxiliary networks; AUM \cite{pleiss2020identifying} achieves this by observing the training dynamics to similarly come up with a score, and then apply threshold to discard ``noisy'' samples. Other suggested methods involve using a guaranteed clean smaller subsample of the entire data to draw deductions about the dataset and using them to filter out anomalous entities from the  dataset (e.g. \cite{li2019learning,ren2019learning,sukhbaatar2015training,Yuncheng}). We tackle the task constrained on the strict setting by identifying required data points without any access to clean subsample and prior knowledge of the type of noise present in the data. 

Furthermore, recent works in anomaly detection use posterior inferences drawn by training variational autoencoders. These methods propose altercations to the architecture and loss functions of standard variational autoencoder to learn latent space representations of the data. Pol et al. \cite{8999265} described a metric based on autoencoder loss function for the task of anomaly detection. Beggel et al. \cite{10.1007/978-3-030-46150-8_13} identified the said points during training based on the likelihood estimate of the latent variable such that anomalous points are expected to get a lower estimate comparative to regular data points. However, our work primarily focuses on deriving a cleaner subset by identifying mislabeled samples in a considerably high noise setting.

\begin{figure*}[!htb]
    \centering{
    \includegraphics[scale=0.4]{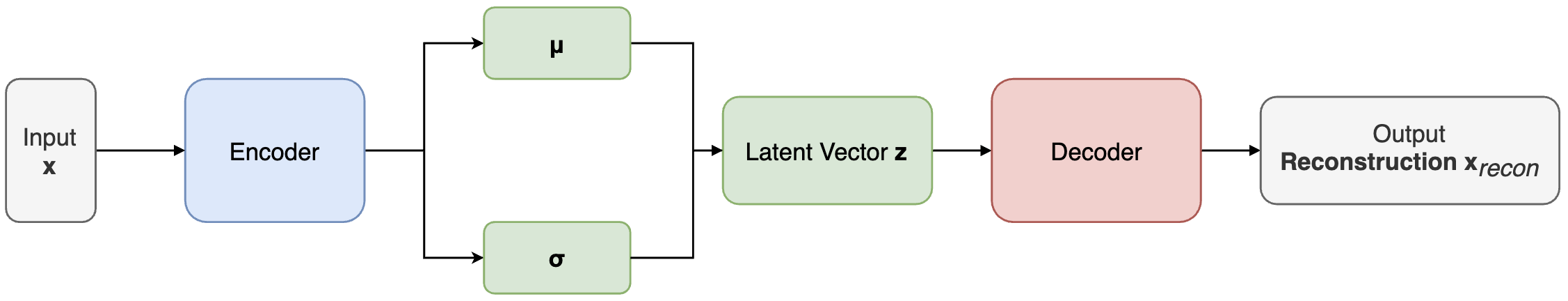}}
    \caption{A standard VAE architecture consists of an encoder model (inference) and a decoder model (generative).  The $\mu$ and $\sigma$ layers encode the mean and standard deviation of the Gaussian prior of the latent variables. The latent vector $z$ is then sampled using the \textit{reparameterization trick} \cite{Kingma2014AutoEncodingVB, pmlr-v32-rezende14}.}
    \label{fig:stdvae}
\end{figure*} 

\begin{figure*}[h]
    \centering{
    \includegraphics[scale=0.4]{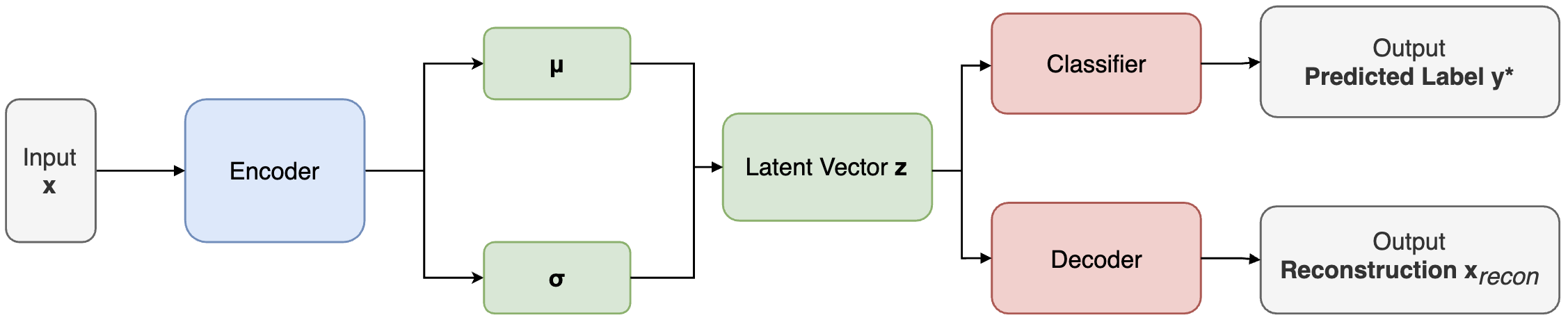}}
    \caption{An architecture of a  supervised VAE: \modelname\ extends standard VAE by adding an auxiliary discriminative network.}
    \label{fig:supvae}
\end{figure*}

\section{Methodology}\label{methods}

We denote the dataset by $D=\{x_{i},y_{i}\}^{N}_{i=1}$ for our task. It consists of two types of samples -- correctly labeled and mislabeled samples. A mislabeled sample is one whose actual label (i.e., ground-truth) does not match the assigned label. We follow a general assumption that such samples hamper the model generalizability. For large-scale datasets, correctly labeled sample could also possibly contribute negatively and might be confusing to the model; but a simplifying assumption is made in this case that it does not hurt model's generalization. We first evaluate our method for the task of noise reduction through identification of mislabeled points. Based on our assumption, we remove such points and evaluate them for the classification task of each dataset using clean test data.

\textbf{Variational Autoencoder:}
It is often employed for efficient approximation of probabilistic inference of the continuous latent variable $z$ given an observed value $x$. The framework of variational autoencoders (VAEs) \cite{Kingma_2019} allows us to explore variations in the data, by learning latent space representation through probabilistic inferences.

For a dataset $D$, standard VAE learns an unsupervised task of stochastic mapping between an observed input ($x$-space) and a continuous latent ($z$-space). Figure \ref{fig:stdvae} shows an overview of a standard VAE architecture. For our approach, we employ a modification to the standard VAE, extending an auxiliary discriminative network that learns y-labels of data.  
  
\begin{figure*}[h]
    \centering{
    \includegraphics[scale=0.45]{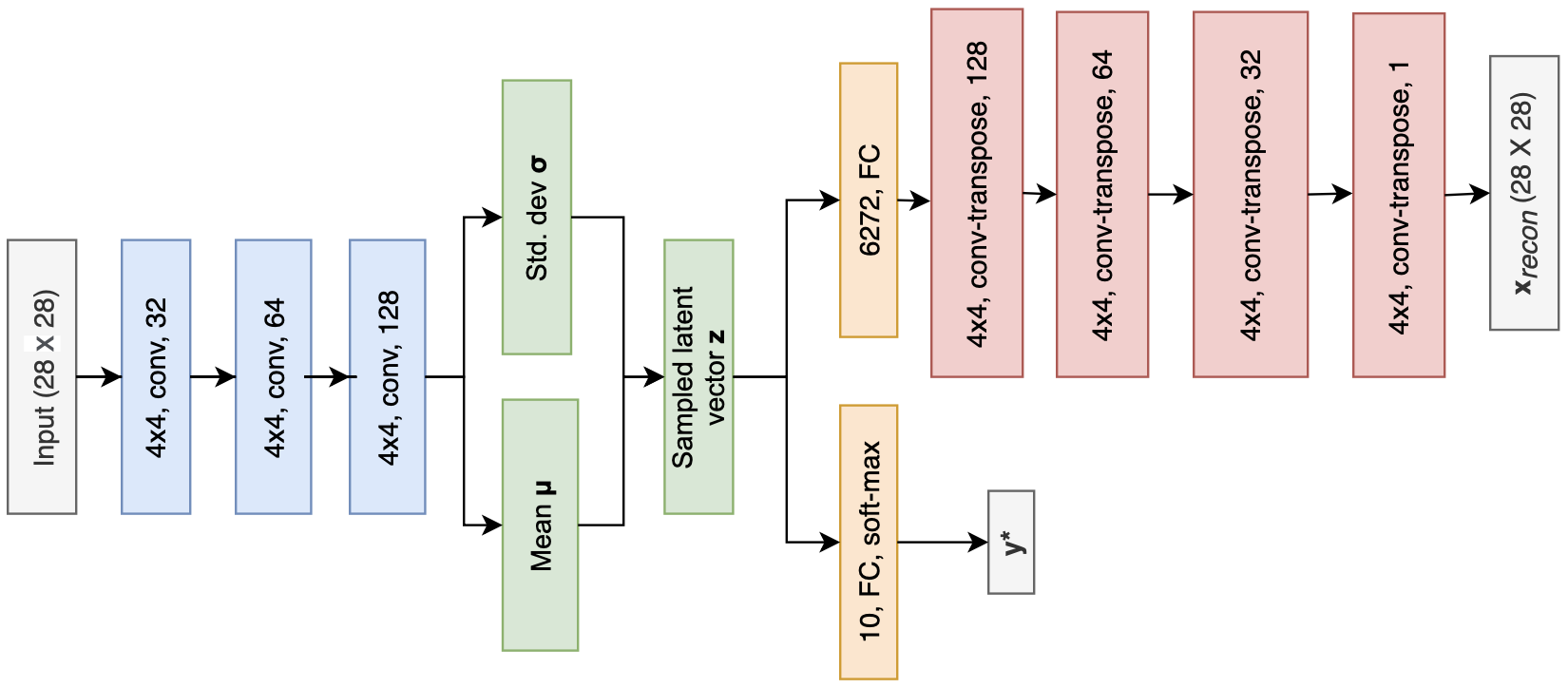}}
    \caption{Detailed architecture of the supervised VAE used for learning latent representations for MNIST and FashionMNIST datasets.}
    \label{fig:mnistsupvae}
\end{figure*}

\begin{figure*}[h]
    \centering{
    \includegraphics[scale=0.45]{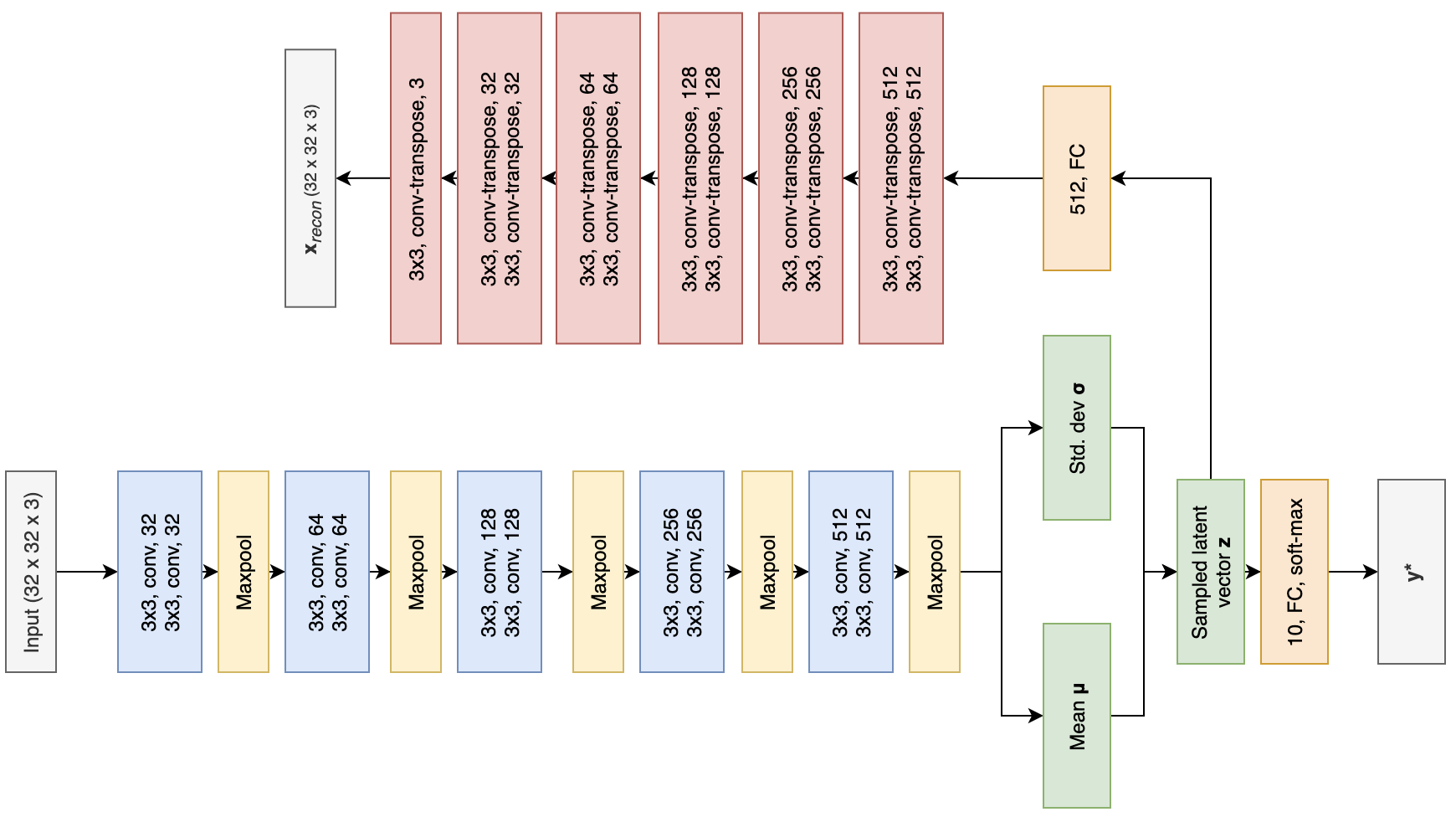}}
    \caption{Detailed architecture of the supervised VAE used for learning latent representations for CIFAR-10/100 datasets.}
    \label{fig:cifarsupvae}
\end{figure*}

\subsection{Proposed Architecture} 
Recent studies  \cite{Berkhahn2019AugmentingVA, 10.5555/2969033.2969226}  suggested supervised and semi-supervised architecture for VAE to learn  latent space representations,  utilizing y-labels corresponding to each observation. The authors showed that the availability of labels improves the VAE encoder inference. Therefore, for our experiment, we implement a supervised VAE model, called \modelname. Figure \ref{fig:supvae} shows an overview of the \modelname\ model. The architecture is largely inspired by the latent-feature discriminative model (M1) proposed by \cite{Kingma2014SemisupervisedLW}. However, unlike the M1 model, where the classifier is trained after learning feature representation, we jointly train for both classification and feature representation.

Specifically, for an input observation $x$, the goal of \modelname\ is to train a network given by $p(x,\, z,\,y) = p(z)\,p(x\,|\,z)\,p(y\,|\,z)$, where $p(z)$ is the prior distribution over latent variables $z$, $p(x\,|\,z)$ is the likelihood function for decoder that generates $x$ given latent variable $z$, and $p(y\,|\,z)$ is the likelihood function for classifier which predicts the label $y$ corresponding to input $x$ given latent variable $z$. This consists of posterior inference network $q_{\phi}(z\,|\,x)$ to infer latent variable $z$, a generative network $p_{\theta}(x|z)$ to reconstruct input $x$, and a discriminative network $q_{\phi}(y\,|\,z)$ to predict label $y$. Therefore, for an input $x_{i}$ belonging to $D$, \modelname\  returns a pair of reconstructed input image and output label i.e. $(x_{i_{recon}},\, y_{i}*)$. 

\subsection{Model Objective} 
We denote $y$ to be the label of the input datapoint $x$.  We describe the following loss functions for optimizing model and variational parameters denoted by $\theta$ and $\phi$, respectively as follows:
\begin{itemize}
    \item \textbf{Autoencoder Loss} is given by evidence lower bound (ELBO) loss function. Here, $z$ denotes the latent variables.
        \begin{equation}
        \begin{split}
            L_{ELBO} =
        &E_{q_{\phi}(z\,|\,x)}[\,log p_{\theta}(x\,|\,z)\,] 
         \,-\,   KL [\,q_{\phi}(z\,|\,x)\,||\,p_{\theta}(z)\,]
        \end{split}
        \end{equation}
        
        The ELBO loss has two terms - first term corresponds to the reconstruction loss, and the second is a regularization term given by Kullback-Leibler divergence.
    \vspace{1mm}
    \item \textbf{Classifier Loss} is given by categorical cross-entropy loss function.
    \begin{equation}
        L_{cl} = \sum\limits_{i}^{}{y_i \cdot logp(y_i)}
    \end{equation}
         
\end{itemize}

Our \modelname\ model is trained to minimize the training objective given by, 
\begin{equation}
        \label{eq:totalLoss}
        L = L_{ELBO} + L_{cl}
\end{equation}

\begin{figure*}[h]
    \centering{\includegraphics[scale=0.31]{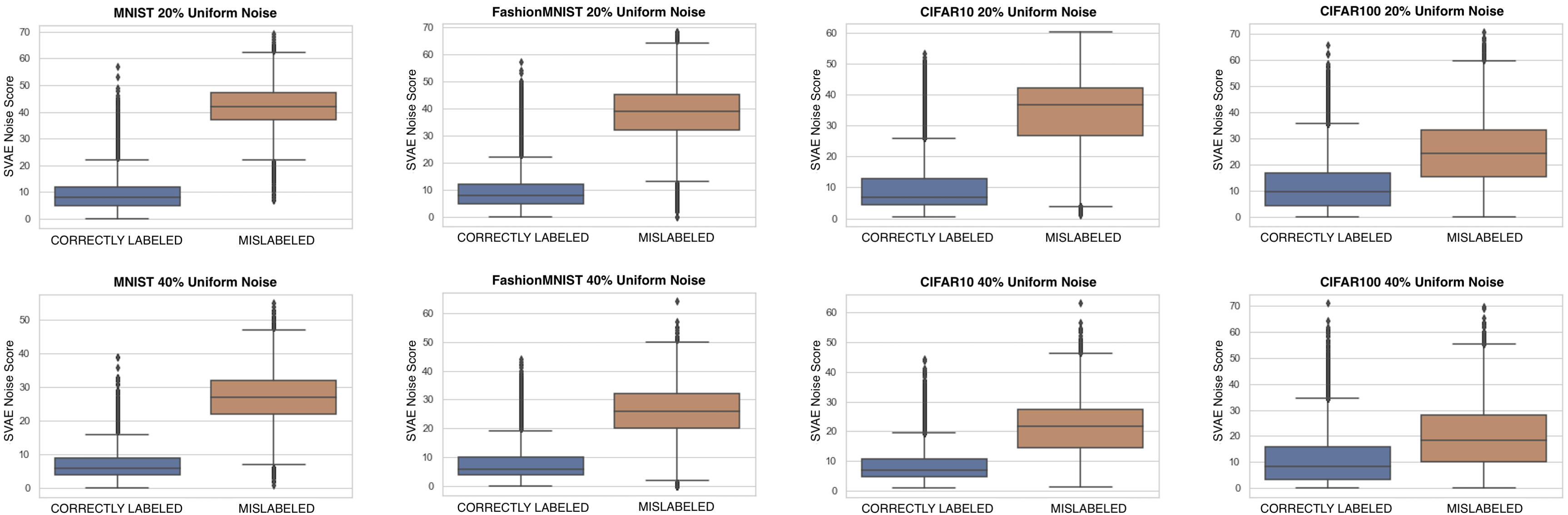}}
    \caption{Illustration for comparing noise scores of correctly labeled and mislabelled samples for MNIST, FashionMNIST, CIFAR-10, and CIFAR-100. The \modelname\ latent space dimension for each dataset has a size of 100 (hence, \textit{noise score} $<= 100$). The box plots distinguish between the values computed for correctly and mislabeled data points as we observe mislabelled samples getting higher noise scores comparative to correctly labeled samples.}
    \label{fig:boxplots}
\end{figure*}

\subsection{Computing Noise Scores} 
We train \modelname\  to learn data distributions in a latent space. Latent space representation or embedding for an input observation is obtained via an inference network (encoder) of \modelname. The proposed metric for the quality of datapoint \textit{noise score} exploits the distribution of latent variables of the data points belonging to the same class as that of the data point.  The intuition behind the proposed method is that data points belonging to the same class should have similar latent representations. Inspired by a univariate outlier detection approach \cite{LeysC} to work on a multivariate setting of latent vectors, we define \textit{noise score} as the number of outlying latent variables differing from the observations belonging to the same class. Algorithm \ref{fig:algo_noise_lvl} gives the pseudo-code for computing noise score for each data point of a dataset $P$ using median and median absolute deviation statistic on a multivariate distribution (i.e., over latent vectors). A constant $\alpha$ is a hyperparameter which adjusts the stringency of outlier detection on each latent variable. We  set a default value of $\alpha$ as $1.5$. Upon obtaining these noise scores, following our prior assumption that mislabelled points hurt the model's performance, we set a hyperparameter threshold based on noise scores distribution to dump points with high scores.

\begin{algorithm}
\caption{ Pseudo-code for computing Noise Score}
\label{fig:algo_noise_lvl}
\begin{algorithmic} 
\REQUIRE Encoder function $f$, Dataset $P$, Constant $\alpha = 1.5$ 
\STATE Group the dataset $P=\{x_{i},y_{i}\}^{N}_{i=1}$ by the $y-$labels. Obtaining a list of disjoint sets $S = [S_1, S_2,..., S_c]$, such that, $P =  S_1 \cup S_2  ... \cup S_c$
\STATE $NScore \leftarrow \phi $
\FORALL{$S_i$ in $S$}
    \STATE $X_i \leftarrow \{x_j \; | \; (x_j,\, y_j) \in S_i \}$
    \STATE $XEnc \leftarrow f(X_i)$
    \STATE $m \leftarrow Median(XEnc, \; axis = 0)$ 
    \STATE $M \leftarrow MedianAbsoluteDeviation(XEnc, \; axis = 0)$
    \FORALL{$latent\_vector_x$ in $XEnc$}
        \STATE $score \leftarrow 0$
        \FORALL{$x_j$ in $latent\_vector_x$}
            \IF{$x_j > m_j + \alpha*M_j$}
                \STATE $score \leftarrow score + 1$
            \ENDIF
        \ENDFOR
        \STATE $NScore[x] \leftarrow score$
    \ENDFOR
\ENDFOR
\RETURN $NScore$
\end{algorithmic}
\end{algorithm}

\begin{table}[!t]\centering
\caption{Dataset Statistics.}\label{tab:dataset}
\begin{tabular}{@{}lll@{}}
\toprule
Dataset & Size (Train/Test, (Height $\times$ Width) & \#Classes \\ \midrule
MNIST            & 60000/10000, (28 $\times$ 28)                      & 10                 \\
FashionMNIST    & 60000/10000, (28 $\times$ 28)                      & 10                 \\
CIFAR10          & 50000/10000, (32 $\times$ 32)                      & 10                 \\
CIFAR100         & 50000/10000, (32 $\times$ 32)                      & 100                \\\midrule
\bottomrule
\end{tabular}
\end{table}

\section{Competing Methods} \label{comp_methods}
\textbf{Standard:} This is a  baseline for robust training of classification model. For this, we do naive training on the complete training dataset.

\textbf{LabelFix:} Muller and Merkert \cite{Muller_2019} scored data points by the probability of an input observation to be assigned the original label. A hyperparameter threshold is then used to identify mislabeled data points by discarding those with low probability. We took the  publicly-available implementation\footnote{\url{https://github.com/mueller91/labelfix}.} to identify mislabeled data points. We then perform robust training on the filtered subset with the same model setting as ours for the purpose of comparison.

\textbf{Entropy Training Dynamics (Entropy TD):}  For this baseline, we differentiate samples based on training dynamics of entropy metric. The Shannon entropy measure of a data point $x$ at epoch $t$ is given by:
\begin{equation*}
    E^{(t)}(x,y) = \sum_i p_i^{(t)} \cdot log(p_i^{(t)})
\end{equation*}
where $p_i^{(t)}$ is an output of the softmax layer at epoch $t$, which corresponds to the probability of $x$ belonging to the class $i$. 

We take account of training dynamics by averaging entropy over all training epochs. This is given by: 
\begin{equation*}
    E(x,y) = \frac{1}{T} \cdot \sum\limits_{t = 1}^{T}E^{(t)}(x,y)
\end{equation*}
Thresholding is applied to the above metric to classify points with a high value of averaged entropy as mislabeled to obtain the filtered dataset used for training the classification model.

\textbf{Oracle: } This serves as an upper bound for the model performance by training on all guaranteed clean samples of the training dataset.


\begin{table*}[!htp]\centering
\caption{Results of mislabel identification on \textbf{MNIST} and \textbf{FashionMNIST} dataset. We present the benchmarks for the following baselines --  Entropy TD, LabelFix and \modelname. We skip standard baseline and oracle as they are undefined for this task. We highlight the best performing method in bold.}\label{tab:mnist_mislabel}
\scalebox{0.89}{\begin{tabular}{lrrrrrrrrrrrrrr}\toprule
\multicolumn{2}{c}{Dataset} &\multicolumn{3}{c}{MNIST} &\multicolumn{3}{c}{MNIST} &\multicolumn{3}{c}{FashionMNIST} &\multicolumn{3}{c}{FashionMNIST} \\\cmidrule{1-14}
\multicolumn{2}{c}{Noise Type} &\multicolumn{3}{c}{Uniform Noise} &\multicolumn{3}{c}{Systematic Noise} &\multicolumn{3}{c}{Uniform Noise} &\multicolumn{3}{c}{Systematic Noise} \\\cmidrule{1-14}
Noise \% &Method &Precision &Recall &Accuracy &Precision &Recall &Accuracy &Precision &Recall &Accuracy &Precision &Recall &Accuracy \\\cmidrule{1-14}
\multirow{3}{*}{0.1} &Entropy TD &0.9 &0.9 &82.3 &0.89 &0.95 &85.2 &0.92 &0.92 &83.2 &0.91 &0.91 &82.7 \\\cmidrule{2-14}
&LabelFix &1 &1 &99.3 &0.99 &0.99 &98.2 &0.99 &0.99 &98.1 &0.96 &0.96 &94.5 \\\cmidrule{2-14}
&\modelname &0.99 &0.99 &99.1 &\textbf{0.99} &\textbf{0.99} &\textbf{98.8} &0.98 &0.98 &96.7 &\textbf{0.99} &\textbf{0.98} &\textbf{97.8} \\\cmidrule{1-14}
\multirow{3}{*}{0.2} &Entropy TD &0.81 &0.82 &75.1 &0.81 &0.91 &80.2 &0.8 &0.9 &75.2 &0.73 &0.81 &0.77 \\\cmidrule{2-14}
&LabelFix &1 &0.99 &98.9 &0.99 &0.99 &97.9 &0.99 &0.98 &97.3 &0.93 &0.93 &89.3 \\\cmidrule{2-14}
&\modelname &0.99 &\textbf{0.99} &98.3 &\textbf{0.99} &\textbf{0.99} &\textbf{98.2} &0.97 &\textbf{0.98} &95.9 &\textbf{0.97} &\textbf{0.96} &\textbf{94.1} \\\cmidrule{1-14}
\multirow{3}{*}{0.3} &Entropy TD &0.73 &0.93 &70.1 &0.72 &0.8 &74.3 &0.7 &0.91 &67.3 &0.7 &0.81 &62.7 \\\cmidrule{2-14}
&LabelFix &0.99 &0.99 &97.8 &0.97 &0.95 &94.5 &0.96 &0.96 &94.9 &0.88 &0.88 &84.2 \\\cmidrule{2-14}
&\modelname &\textbf{0.99} &\textbf{0.99} &\textbf{98.2} &\textbf{0.99} &\textbf{0.98} &\textbf{97.5} &\textbf{0.98} &0.95 &94.7 &\textbf{0.93} &\textbf{0.92} &\textbf{89.5} \\\cmidrule{1-14}
\multirow{3}{*}{0.4} &Entropy TD &0.71 &0.94 &65.2 &0.7 &0.8 &62.8 &0.61 &0.91 &59.4 &0.6 &0.8 &57.3 \\\cmidrule{2-14}
&LabelFix &0.97 &0.98 &96.1 &0.93 &0.93 &91.7 &0.93 &0.93 &94.1 &0.79 &0.81 &76.7 \\\cmidrule{2-14}
&\modelname &\textbf{0.97} &0.95 &95.5 &\textbf{0.97} &\textbf{0.93} &\textbf{94.1} &\textbf{0.93} &\textbf{0.94} &91.5 &\textbf{0.87} &\textbf{0.84} &\textbf{83.1} \\\midrule
\bottomrule
\end{tabular}}
\end{table*}

\begin{table*}[!htp]\centering
\caption{Results of mislabel identification on \textbf{CIFAR10} and \textbf{CIFAR100} dataset. We present the benchmarks for the following baselines --  Entropy TD, LabelFix and \modelname. We skip standard baseline and oracle as they are undefined for this task. We highlight the best performing method in bold.}\label{tab:cifar_mislabel}
\scalebox{0.89}{
\begin{tabular}{lrrrrrrrrrrrrrr}\toprule
\multicolumn{2}{c}{Dataset} &\multicolumn{3}{c}{CIFAR10} &\multicolumn{3}{c}{CIFAR10} &\multicolumn{3}{c}{CIFAR100} &\multicolumn{3}{c}{CIFAR100} \\\cmidrule{1-14}
\multicolumn{2}{c}{Noise Type} &\multicolumn{3}{c}{Uniform Noise} &\multicolumn{3}{c}{Systematic Noise} &\multicolumn{3}{c}{Uniform Noise} &\multicolumn{3}{c}{Systematic Noise} \\\cmidrule{1-14}
Noise \% &Method &Precision &Recall &Accuracy &Precision &Recall &Accuracy &Precision &Recall &Accuracy &Precision &Recall &Accuracy \\\cmidrule{1-14}
\multirow{3}{*}{20\%} &Entropy TD &0.92 &0.69 &71.2 &0.81 &0.91 &76.2 &0.86 &0.86 &77.4 &0.8 &0.9 &75.4 \\\cmidrule{2-14}
&LabelFix &0.94 &0.96 &92.7 &0.9 &0.92 &86.2 &0.95 &0.95 &92.4 &0.85 &0.85 &75.16 \\\cmidrule{2-14}
&\modelname &0.93 &\textbf{0.96} &90.7 &\textbf{0.93} &0.8 &82.4 &0.9 &\textbf{0.95} &85.1 &\textbf{0.9} &0.75 &72.2 \\\cmidrule{1-14}
\multirow{3}{*}{40\%} &Entropy TD &0.8 &0.8 &75.1 &0.79 &0.79 &74.2 &0.68 &0.9 &68.3 &0.6 &0.9 &58.2 \\\cmidrule{2-14}
&LabelFix &0.91 &0.91 &87.9 &0.67 &0.67 &58.9 &0.9 &0.9 &87.4 &0.63 &0.63 &55.73 \\\cmidrule{2-14}
&\modelname &0.85 &\textbf{0.91} &84.8 &\textbf{0.82} &0.6 &69.2 &0.84 &\textbf{0.9} &83.5 &\textbf{0.66} &0.75 &\textbf{61.2} \\\midrule
\bottomrule
\end{tabular}}
\end{table*}

\section{Experiments}
We conduct our experiments on the following datasets: \textbf{MNIST}, \textbf{FashionMNIST}, \textbf{CIFAR10}, and \textbf{CIFAR100} as shown in Table \ref{tab:dataset}. Since these datasets are well known, we assume them to be correctly labeled i.e., original labels as the \textit{true} labels. We add noise to the training set using two noise models -- (i) {\bf uniform noise model} adds mislabeling by assigning a random label such that each class has an equal probability of getting mislabeled to any other class, and (ii) {\bf systematic noise model} wherein data points belonging to the same true class get assigned the same incorrect label. The latter simulates a scenario where one class has a predisposition to be mistaken for another class, e.g., handwritten digit 8 often gets confused with the digit 3. We describe the mislabeling of the systematic noise model for a dataset with $L$ classes given by $\{0,...,(L\, -\, 1)\}$ by assigning random data points with the initial true label as $l$ to class $(l \,+\, 1)\%L$.

\subsection{Implementation Details} We implement two different \modelname\ networks for our experiments. Detailed architecture is illustrated in Figure \ref{fig:mnistsupvae} used for MNIST and FashionMNIST, and Figure \ref{fig:cifarsupvae} used for CIFAR-10/100. The dimension of 100 is used for latent space. The encoder design implemented for CIFAR (in Figure \ref{fig:cifarsupvae}) has been inspired by VGG-like networks \cite{simonyan2015deep}, and the decoder design uses transposed convolutional layer to reconstruct the input observation. \modelname\ is trained on the noisy training dataset $D$.  To prevent overtraining on the noisy training dataset, we apply early termination on the validation loss given by Equation \ref{eq:totalLoss}. Upon training, the encoder (or inference) network is then used to compute the required \textit{noise scores} for each data point through Algorithm \ref{fig:algo_noise_lvl} on dataset $D$. Figure \ref{fig:boxplots} presents the box plots to discern between the obtained noise scores for correctly labeled and mislabeled points. A hyperparameter threshold is deduced based on the distribution of noise scores of $D$ to filter out points with high noise scores as mislabelled/hard data points.

\begin{table*}[!htp]\centering
\caption{Robust training performance on \textbf{MNIST} and \textbf{FashionMNIST} dataset. We present the baselines over all the methods. Oracle method serves as the upper bound of models performance. We highlight the best performing method in bold.}\label{tab:mnist_train}
\begin{tabular}{l|r|rrrr|r|rrrr}\toprule
\multicolumn{11}{c}{\textbf{Uniform Noise}} \\\cmidrule{1-11}
{\bf Dataset} &\multicolumn{5}{c}{\bf MNIST (5\% subsampled)} &\multicolumn{5}{c}{\bf FashionMNIST} \\\cmidrule{1-11}
Noise \% &Oracle &Standard &\modelname &LabelFix &Entropy TD &Oracle &Standard &\modelname &LabelFix &Entropy TD \\\cmidrule{1-11}
10\% &95.4 &94.3 &\textbf{96.2} &95.6 &94.4 &90.1 &89.5 &\textbf{89.7} &\textbf{89.7} &88.5 \\\cmidrule{1-11}
20\% &94.5 &92.3 &\textbf{95.12} &94.9 &93.5 &89.8 &88.9 &89.4 &\textbf{89.6} &87.4 \\\cmidrule{1-11}
30\% &94.9 &91.9 &\textbf{94.3} &93.9 &92.5 &89.8 &88.1 &\textbf{88.5} &88.3 &87.2 \\\cmidrule{1-11}
40\% &94.5 &90.3 &93.3 &\textbf{93.6} &91.3 &89.2 &86.5 &87.4 &\textbf{87.9} &86.3 \\\cmidrule{1-11}\midrule
\multicolumn{11}{c}{\textbf{Systematic Noise}} \\\cmidrule{1-11}
{\bf Dataset} &\multicolumn{5}{c}{\bf MNIST (5\% subsampled)} &\multicolumn{5}{c}{\bf FashionMNIST} \\\cmidrule{1-11}
Noise \% &Oracle &Standard &\modelname &LabelFix &Entropy TD &Oracle &Standard &\modelname &LabelFix &Entropy TD \\\cmidrule{1-11}
10\% &95.7 &92.6 &\textbf{95.6} &95.3 &93.7 &90.1 &89.3 &\textbf{89.8} &89.4 &89.1 \\\cmidrule{1-11}
20\% &94.8 &91.2 &\textbf{94.8} &94.6 &90.6 &90.2 &88.9 &\textbf{89.1} &88.7 &87.5 \\\cmidrule{1-11}
30\% &94.3 &91.7 &\textbf{94.6} &93.7 &89.2 &89.1 &85.3 &\textbf{87.5} &86.2 &84.6 \\\cmidrule{1-11}
40\% &94.2 &88.8 &\textbf{93.1} &91.2 &88.2 &88.8 &74.5 &\textbf{85.1} &82.4 &78.6 \\\midrule
\bottomrule
\end{tabular}
\end{table*}


\begin{table*}[!htp]\centering
\caption{Robust training performance on \textbf{CIFAR10} and \textbf{CIFAR100} dataset. We present the baselines over all the methods. Oracle method serves as the upper bound of models performance. We highlight the best performing method in bold.}\label{tab:cifar_train}
\begin{tabular}{l|r|rrrr|r|rrrr}\toprule
\multicolumn{11}{c}{\textbf{Uniform Noise}} \\\cmidrule{1-11}
\textbf{Dataset} &\multicolumn{5}{c|}{\textbf{CIFAR10} } &\multicolumn{5}{c}{\textbf{CIFAR100}} \\\cmidrule{1-11}
Noise \% &Oracle &Standard &\modelname &LabelFix &Entropy TD &Oracle &Standard &\modelname &LabelFix &Entropy TD \\\cmidrule{1-11}
20\% &89.4 &72.6 &82.5 &\textbf{84.2} &77.5 &61.1 &47.7 &50.7 &\textbf{52.3} &51.1 \\\cmidrule{1-11}
40\% &85.5 &57.5 &71.2 &\textbf{76.5} &64.8 &57.5 &39.7 &40.1 &\textbf{43.1} &39.6 \\\cmidrule{1-11}\midrule
\multicolumn{11}{c}{\textbf{Systematic Noise}} \\\cmidrule{1-11}
{\bf Dataset} &\multicolumn{5}{c|}{\bf CIFAR10 } &\multicolumn{5}{c}{\bf CIFAR100} \\\cmidrule{1-11}
Noise \% &Oracle &Standard &\modelname &LabelFix &Entropy TD &Oracle &Standard &\modelname &LabelFix &Entropy TD \\\cmidrule{1-11}
20\% &89.5 &69.1 &\textbf{81.5} &81.2 &72 &61.7 &52.2 &\textbf{53.1} &51.4 &52.4 \\\cmidrule{1-11}
40\% &85.8 &51.9 &\textbf{68.7} &56.2 &53.74 &58.1 &38.9 &\textbf{39.1} &33.9 &38.2 \\\midrule
\bottomrule
\end{tabular}
\end{table*}

\subsection{Identification of Mislabeled Data}

For the first task of our pipeline i.e., filtering out correctly labeled points from the dataset, we use precision, recall, and accuracy as the evaluation metrics. For a resultant filtered subset, precision indicates the percentage of clean data points among all the filtered data points, measuring the quality of the filtered dataset. Recall measures the percentage of clean data points in the filtered subset among all clean data points of the original dataset. Lastly, accuracy measures the correctness of the method's segregation. 


Tables \ref{tab:mnist_mislabel} and \ref{tab:cifar_mislabel} present results for the task of mislabel identification. 
%
Our method achieves high precision, recall, and accuracy (consistently $>= 0.94$) for MNIST and against both noise models for varying amounts of corruption up to $40\%$. In general, for all methods, systematic noise is observed to be harder to filter compared to uniform noise.   This task becomes more challenging for CIFAR datasets. We note CIFAR100 with high noise as the most challenging setting, due to it being a more complex dataset compared to others. In particular, our method shows improvement to the results of baselines against systematic noise for MNIST, FashionMNIST, and CIFAR10.

\subsection{Robust Training}

We discard the identified mislabeled samples and train for the classification task of these datasets. We evaluate models performance on a \textit{clean} test set. We compare our results against all the methods described in Section \ref{comp_methods}. We implement RESNET32 \cite{he2016deep}  for CIFAR10/100 dataset and 2-layer vanilla CNN network for MNIST and FashionMNIST datasets. Since MNIST is a relatively simpler dataset to model, in order to distinguish all results, we train the model on a  5\% sample of the training dataset. Tables \ref{tab:mnist_train} and \ref{tab:cifar_train} present results for robust training of classification model. In all cases, We note significant improvements against the standard baseline. Moreover, in case of MNIST, the accuracy of our method exceeds that of oracle for 10\% and 20\% noise. Additionally, for a more challenging setting of CIFAR-100 as shown in Table \ref{tab:cifar_train}, we only see slight improvements to the standard baselines. These results are indicative that the quality of datasets could bring more improvements to having more data.

\section{Conclusion}
In our paper, we introduced \modelname, a VAE based architecture with an auxiliary discriminative network. \modelname\ combines the data labels to better learn the latent space representations improving the models encoder inference. We also introduced a novel \textit{noise score} metric, which exploits the variations in the latent representation of data points belonging to the same class. This measure is further used to order data points on the quality and segregate noisy or mislabeled ones.  We tested our technique for varying amounts of label error using two noise models -- uniform and systematic. We showed that our method is successfully able to distinguish mislabeled and correctly labeled points. Note that our method does not use any prior information on the type of noise present in data. Due to the prevalence of large parameterized deep neural networks, they end up learning this noise, which hurts the generalization ability of the model. Hence, we remove the identified mislabeled data points and observe significant improvements in the performance of the classification model. Furthermore, since our algorithm relies on an inference network, we believe that exploring other autoencoder architectures could provide further improvements to existing results. In particular, for large datasets, where noise invariably creeps up in real-world settings, we believe our proposed pipeline will be useful in identifying and removing low-quality data points.

\bibliography{bibliography}

\end{document}